\documentclass[acmsmall,screen,nonacm]{acmart}

\usepackage{amsmath}
\usepackage{amsmath,bm}
\usepackage{multirow}
\usepackage{color}
\usepackage{amsthm}
\usepackage{graphicx}
\usepackage{subcaption}
\usepackage{algorithm}
\usepackage{algorithmic}
\usepackage{soul}
\usepackage{hyperref}
\usepackage{makecell}
\usepackage{bbm}
\usepackage{tabularx}

\AtBeginDocument{%
  }

\begin{document}

\title{Exploring Fairness in Educational Data Mining in the Context of the Right to be Forgotten}

\author{Wei Qian}
\authornote{Both authors contributed equally to this research.}
\email{wqi@iastate.edu}
\affiliation{
  \institution{Iowa State University}
  \country{USA}
}
\author{Aobo Chen}
\authornotemark[1]
\email{aobochen@iastate.edu}
\affiliation{
  \institution{Iowa State University}
  \country{USA}
}
\author{Chenxu Zhao}
\email{cxzhao@iastate.edu}
\affiliation{
  \institution{Iowa State University}
  \country{USA}
}
\author{Yangyi Li}
\email{liyangyi@iastate.edu}
\affiliation{
  \institution{Iowa State University}
  \country{USA}
}
\author{Mengdi Huai}
\email{mdhuai@iastate.edu}
\affiliation{
  \institution{Iowa State University}
  \country{USA}
}

\renewcommand{\shortauthors}{Wei Qian, Aobo Chen, Chenxu Zhao, Yangyi Li, and Mengdi Huai}

\begin{abstract}
In education data mining (EDM) communities, machine learning has achieved remarkable success in discovering patterns and structures to tackle educational challenges. Notably, fairness and algorithmic bias have gained attention in learning analytics of EDM. With the increasing demand for the right to be forgotten, there is a growing need for machine learning models to forget sensitive data and its impact, particularly within the realm of EDM. The paradigm of selective forgetting, also known as machine unlearning, has been extensively studied to address this need by eliminating the influence of specific data from a pre-trained model without complete retraining. However, existing research assumes that interactive data removal operations are conducted in secure and reliable environments, neglecting potential malicious unlearning requests to undermine the fairness of machine learning systems. In this paper, we introduce a novel class of selective forgetting attacks designed to compromise the fairness of learning models while maintaining their predictive accuracy, thereby preventing the model owner from detecting the degradation in model performance. Additionally, we propose an innovative optimization framework for selective forgetting attacks, capable of generating malicious unlearning requests across various attack scenarios. We validate the effectiveness of our proposed selective forgetting attacks on fairness through extensive experiments using diverse EDM datasets.

\end{abstract}



\keywords{selective forgetting, educational data mining, fairness, the right to be forgotten}


\maketitle

\section{Introduction}
\label{sec:intro}

Over the years, extensive research has been focusing on educational data mining (EDM). Student data, which is a critical component in EDM research, can contain personal information, such as age and gender, as well as academic performance and activity data from online learning systems~\cite{hutt2023right}. By offering valuable insights into student learning, EDM supports the development of more effective educational practices and policies, ultimately improving student outcomes. One of the most popular techniques in the previous works is incorporating machine learning techniques, which has achieved remarkable success in discovering intricate structures within educational datasets. However, in recent years, concerns about the fairness of deploying algorithmic decision-making in the educational context have emerged~\cite{kizilcec2022algorithmic, vasquez2022faired, baker2022algorithmic, holstein2022equity}. Particularly, machine learning models can produce biased and unfair outcomes for certain student groups, significantly affecting their educational opportunities and achievements.

Given that the data empowering EDM research can often contain personally identifiable and other sensitive information, there has been increased attention to privacy protection in recent years~\cite{student2017, reidenberg2018achieving}. Additionally,  privacy legislation such as the California Consumer Privacy Act \cite{pardau_california_nodate} and the former Right to be Forgotten \cite{ginart2019making} has granted users the right to erase the impact of their sensitive information from the trained models to protect their privacy. One approach to protecting users' privacy involves enabling the trained machine learning model to entirely forget and remove the influence of the specific data points to be erased, without compromising the contributions of other data points. Moreover, the presence of poor-quality data, such as noise or outliers, can significantly degrade the performance of trained models. Therefore, it is necessary for an EDM system to remove such data points to regain utility. The simplest way to achieve such forgetting demands is to train a new model on all data, excluding the removed portion. However, this approach is generally impractical due to the tremendous computational resources it consumes. Hence, to efficiently remove data along with their impact on the pre-trained model, a novel field in machine learning privacy protection has developed, known as \emph{selective forgetting} (\emph{machine unlearning})~\cite{cao2015towards, guo2019certified, golatkar2020eternal, bourtoul2021, qian2022patient, yan2022arcane, warnecke2023machine, liu2024threats, wang2023inductive, liu2023muter}.

Considerable exploration has been made into selective forgetting, aiming to fulfill data deletion requirements while minimizing computational expenses. Existing methods are primarily divided into two categories: exact unlearning~\cite{bourtoul2021,yan2022arcane, qian2022patient} and approximate unlearning~\cite{guo2019certified, golatkar2020eternal,shibata2021learning,mehta2022deep, warnecke2023machine}. The exact unlearning algorithms are designed to reduce the time complexity during the retraining phase. For example, SISA~\cite{bourtoul2021} divides the dataset into smaller shards, with each shard being used to train a separate shard model.  This approach means that only the shard models containing the samples to be forgotten need to be retrained during the unlearning process, thereby reducing the overall computation time. In contrast, the approximate training approaches seek to achieve a close approximation through post-processing. For instance,~\cite{warnecke2021machine} uses influence functions to measure the impact of training points on a learning model's predictions.

However, existing studies on selective forgetting are mostly about creating effective unlearning algorithms to remove the data from models, assuming that interactive data removal operations are conducted in secure and reliable environments and neglecting potential malicious unlearning requests to undermine the security of machine learning systems. Although there exist recent works investigating the security impact of unlearning~\cite{qian2023towards, zhao2024static, liu2024backdoor}, they primarily focus on how malicious unlearning requests can reduce the prediction accuracy of machine learning models. Nevertheless, the impact of unlearning requests on the fairness of machine learning models still remains unknown. In practice, a motivated attacker could exploit the selective forgetting pipeline to interfere with fairness in EDM. Consider the example where unlearning requests are utilized to protect the privacy of students; maliciously removing records of a particular demographic group during the unlearning process could lead to dataset imbalance and predictive biases~\cite{calmon2017optimized}. On the other hand, current research on unlearning in relation to fairness mainly focuses on achieving fairness and unlearning simultaneously~\cite{oesterling2024fair,zhang2024forgotten,shao2024federated}. Our work, however, investigates the impact of unlearning requests on fairness within EDM communities.

To bridge this gap, we perform a comprehensive study on the vulnerability of fairness on educational data mining systems to malicious unlearning requests during the unlearning process. To be more specific, we propose a new framework of selective forgetting attacks based on the fact that a batch of unlearning requests can be processed such that the unlearned models exhibit increasing disparities among different student groups while still maintaining prediction accuracy. As a result, the attack is difficult to be perceived by the model owner. Moreover, we formulate our fairness loss at the group level and the individual level, which are two commonly studied forms of fairness in machine learning~\cite{dwork2012fairness, hajian2016algorithmic, berk2017convex}. Further, considering practical unlearning scenarios, our proposed framework is able to perform both whole~\cite{koh2017understanding} (i.e., removing entire data samples) and partial~\cite{kim2022efficient, liu2022backdoor, warnecke2021machine} (i.e., removing partial data information) unlearning algorithms. We conduct extensive experiments under various attack scenarios to validate the effectiveness of our selective forgetting attacks. To the best of our knowledge, this study is the first to explore machine unlearning as a strategy for adversarial machine learning, specifically investigating the fairness vulnerability of EDM systems during the unlearning process.
\section{Related Work}
\label{sec:related}




Ensuring fairness in educational data mining is crucial for guaranteeing that the benefits of data-driven educational technologies are equitably distributed among all students, regardless of their demographic characteristics. Existing literature in this field can be categorized into three primary areas: measuring fairness~\cite{gardner2019evaluating,verger2023your,cohausz2024fairness}, understanding the implications of unfairness~\cite{cohausz2023investigating,kizilcec2022algorithmic}, and designing fair models for EDM~\cite{le2023review,quy2021fair}. Common metrics in machine learning for assessing fairness include group fairness and individual fairness. Group fairness aims to ensure the equality of predictive performance across different groups, incorporating measures such as Statistical Parity~\cite{feldman2015certifying}, Equalized Odds~\cite{hardt2016equality}, and Equal Opportunity~\cite{hardt2016equality}. Individual fairness~\cite{dwork2012fairness}, on the other hand, ensures that similar individuals receive similar predictive outcomes. Additionally, several novel fairness metrics have been introduced in EDM to more accurately reflect the subtleties of fairness in educational contexts, including the Absolute Between-ROC Area (ABROCA)~\cite{gardner2019evaluating}, and the Model Absolute Density Distance (MADD)~\cite{verger2023your}. 
Research has also focused on understanding the broader implications of unfairness within EDM. For example, \cite{cohausz2023investigating} shows that demographic features do not increase a model’s performance and argues leaving out demographic features for prediction;~\cite{kizilcec2022algorithmic} identifies several key sources of bias and discrimination in the EDM system. Efforts to design fair models for EDM have led to the development of algorithms specifically tailored to mitigate bias in educational data. For example,~\cite{quy2021fair} proposes a novel clustering method, which can cluster similar samples while preserving cluster fairness. Despite these advancements, there remains a gap in addressing fairness in EDM from the perspective of malicious attacks, which pose significant threats to the integrity and fairness of educational data models.



Selective forgetting, also known as machine unlearning, aims to remove the influence of request deleted data from a well-trained model. There are two main categories: exact unlearning, exemplified by SISA~\cite{bourtoule2021machine}, and approximate unlearning, which includes first-order methods~\cite{warnecke2021machine}, second-order methods~\cite{warnecke2021machine}, and unrolling SGD~\cite{thudi2022unrolling}. This field has seen development across various domains such as healthcare~\cite{zhou2023unified} and education~\cite{hutt2023right}. There is a growing body of research on the potential risks associated with the unlearning phase~\cite{chen2021machine, hu2023duty, di2022hidden, qian2023towards, zhao2024static}. For instance, \cite{chen2021machine} investigates privacy risks by comparing models before and after unlearning.~\cite{hu2023duty} examines potential threats within the context of Machine Learning as a Service (MLaaS).~\cite{di2022hidden} introduces a new attack paradigm involving the addition of carefully crafted points followed by unlearning a subset of these points as a trigger.~\cite{qian2023towards} designs malicious perturbations on request-unlearned data to achieve an attacker’s goals.~\cite{zhao2024static} formulates the unlearning attack in a sequential unlearning setting. Additionally, several studies have focused on the fairness implications of the unlearning process~\cite{shao2024federated, kadhe2023fairsisa, zhang2024forgotten, chen2024fast, oesterling2024fair}.~\cite{zhang2024forgotten} analyzes the effect of random deletion on a model's fairness.~\cite{chen2024fast} utilizes unlearning as a tool for fast model debiasing. Other works explore fairness in various settings, such as in large language models (LLMs)~\cite{kadhe2023fairsisa} and federated learning~\cite{shao2024federated}. In contrast to these works, our paper aims to understand the fairness effects during the unlearning process by framing it as an adversarial problem to comprehend adversary behaviors better.



\section{Preliminary}

In this work, we consider the classification problem within the field of educational data mining. Let $\mathcal{D}=\{z_i = ({x}_{i},y_{i})\}_{i=1}^{n}$ denote the training dataset, where ${x}_{i} \in \mathbb{R}^{D}$ is a sample and $y_{i} \in [C]=\{1,\cdots,C\}$ indicates its corresponding class label. We define a learning algorithm (that is used to train the classifier) as $f(\theta): \mathcal{X} \rightarrow \mathbb{R}^{C}$, where $\mathcal{X} \subseteq \mathbb{R}^{D}$ is the $D$-dimensional input domain, and $\theta \in \Theta$ denotes the set of model weights that parameterize the model. We denote $F(x;\theta)$ as the output logits of the learning model on input ${x}\in \mathcal{X}$.\\

\noindent\textbf{Unlearning.} Note that the goal of machine unlearning is to remove the influence of certain data that need to be forgotten from a pre-trained model. Here, we define an unlearning method $\mathcal{U}$ that takes the pre-trained model $f(\theta^*)$, the original training dataset $\mathcal{D}$, and the data to be forgotten $\mathcal{D}_u \in \mathcal{D}$, to derive an unlearned model $\mathcal{U}(\theta^{*}, \mathcal{D}, \mathcal{D}_u)$. Ideally, the unlearned model is expected to closely resemble the model obtained by retraining from scratch on the remaining data $\mathcal{D} \setminus \mathcal{D}_u$. In the following, we describe some popular unlearning methods. 
\begin{itemize}
    \item \emph{First-order based unlearning method~\cite{warnecke2023machine}.} This method uses a first-order Taylor series to derive the gradient updates. Let ${Z}=\{{z}_{p}\}_{p=1}^{P} \subset \mathcal{D}$ denote a set of targeted training samples and $\tilde{{Z}}=\{\tilde{{z}}_{p}\}_{p=1}^{P}$ for the corresponding unlearned versions, where $\tilde{{z}}_{p}=({x}_{p}-{\delta}_{p},y_{p})$ and ${\delta}_p$ is the unlearning information for ${x}_{p}$. The unlearned model ${\theta}^{u}$ can be obtained by updating the model parameters as $\theta^{u} \leftarrow  \theta^{*}-\tau (\sum_{\tilde{{z}}_{p}\in \tilde{{Z}}} \nabla_{} \ell(\tilde{{z}}_{p}; {\theta}^{*}) -\sum_{{z}_{p} \in {Z}} \nabla_{} \ell({z}_{p};{\theta}^{*}))$, where ${\theta}^{*}$ is a pre-trained model, $\tau$ is a pre-defined unlearning rate, and $\ell$ is a loss function (e.g., cross-entropy).

    \item \emph{Second-order based unlearning method \cite{warnecke2023machine}.} This method applies the inverse Hessian matrix of the second-order derivatives for unlearning. The unlearned model ${\theta}^{u}$ is formulated as ${\theta}^{u} \leftarrow {\theta}^{*}-{H}_{{\theta}^{*}}^{-1} (\sum_{\tilde{{z}}_{p}\in \tilde{{Z}}} \nabla_{{\theta}} \ell(\tilde{{z}}_{p}; {\theta}^{*}) 
    - \sum_{{z}_{p} \in {Z}} \nabla_{{\theta}} \ell({z}_{p};{\theta}^{*}))$, where ${H}_{{\theta}^{*}}^{-1}$ is the inverse Hessian matrix and $\ell$ is a loss function (e.g., cross-entropy).

    \item \emph{Unrolling SGD unlearning method \cite{thudi2022unrolling}.} Unrolling SGD formalizes a single gradient unlearning method by expanding a sequence of stochastic gradient descent (SGD) updates using a Taylor series. To reverse the effect of unlearning data during the SGD training steps and obtain the unlearned model, this method involves adding the gradients of the unlearning data, computed with respect to the initial weights, into the final model weights.

    \item \emph{SISA \cite{bourtoul2021}.} In SISA, the original training dataset is randomly partitioned into several disjoint shards. For each shard, a corresponding shard model is trained using the data from that shard. Subsequently, the final prediction results are obtained through the aggregation of shared models (e.g., via majority voting). Upon receiving the forgotten data, the model provider only needs to retrain the specific shard model that includes the forgotten data within its shard.
\end{itemize}

\noindent\textbf{Fairness.} The fairness in machine learning assumes that the model should not be biased on sensitive attributes, such as gender, age, or race. Each sensitive attribute partitions a population into different groups, including the privileged group and the unprivileged group. In this work, we assess the biases of the model with the concept of Equalized Odds~\cite{hardt2016equality}. Let $S=\{a, b\}$ represent a sensitive attribute, $Y$ denotes the class label, and $\hat{Y}$ signifies the predicted outcome. A model $f(\theta)$ satisfies Equalized Odds with respect to the sensitive attribute $S$ and the label $Y$ if the predicted outcome $\hat{Y}$ and $S$ are independent conditional on $Y$. More formally, for all $y\in \{0, 1\}$,
\begin{align}
    & Pr(\hat{Y}=1 | S=a, Y=y) = Pr(\hat{Y}=1 | S=b, Y=y).
\end{align}
We build our prior approaches by incorporating the unfairness measure into the selective forgetting problem. In the following, we introduce our selective forgetting attacks designed to exacerbate the fairness gap during the unlearning process.
\section{Methodology}
\label{sec:method}
In this section, we begin by presenting the considered threat model. Then, we design a general attack framework to identify effective forgetting attack strategies for exploring fairness in educational data mining. After that, we give more discussions on the proposed selective forgetting attacks.

\subsection{Threat Model}
In selective forgetting attacks, we examine a threat model involving a model owner possessing a well-trained model and an attacker aiming to manipulate the unlearning process of the pre-trained model. The attacker may masquerade as a data provider utilized by the pre-trained model, with the intention of inducing the model owner to execute deleterious unlearning actions, typically driven by privacy concerns or some potential conflicts of interest. Once the model owner complies with the attacker's requests and removes the information associated with the specified data from the well-trained model, the correspondingly unlearned model exhibits biases in its predictions to inputs. We highlight that the attacker only initiates unlearning requests to forget certain data during the unlearning process, lacking the capability to alter training data during the training phase or test data during the inference phase. Moreover, the unlearned model is expected to maintain its predictive performance to ensure that our selective forgetting attacks remain stealthy and inconspicuous.

In this paper, we study both the while-box and the black-box settings. The white-box setting assumes the attacker possesses complete knowledge of the learning system, including the model architecture and parameters. This setting is crucial to exploring the strongest attacking behavior in adversarial machine learning and has been widely adopted in much literature on evasion and poisoning attacks. Conversely, the black-box setting assumes the attacker lacks prior knowledge of the pre-trained model, representing a more realistic scenario in real-world applications.

\subsection{Attack Formulation}

\begin{figure*}[t!]
    \centering
    \includegraphics[width=1.0\linewidth]{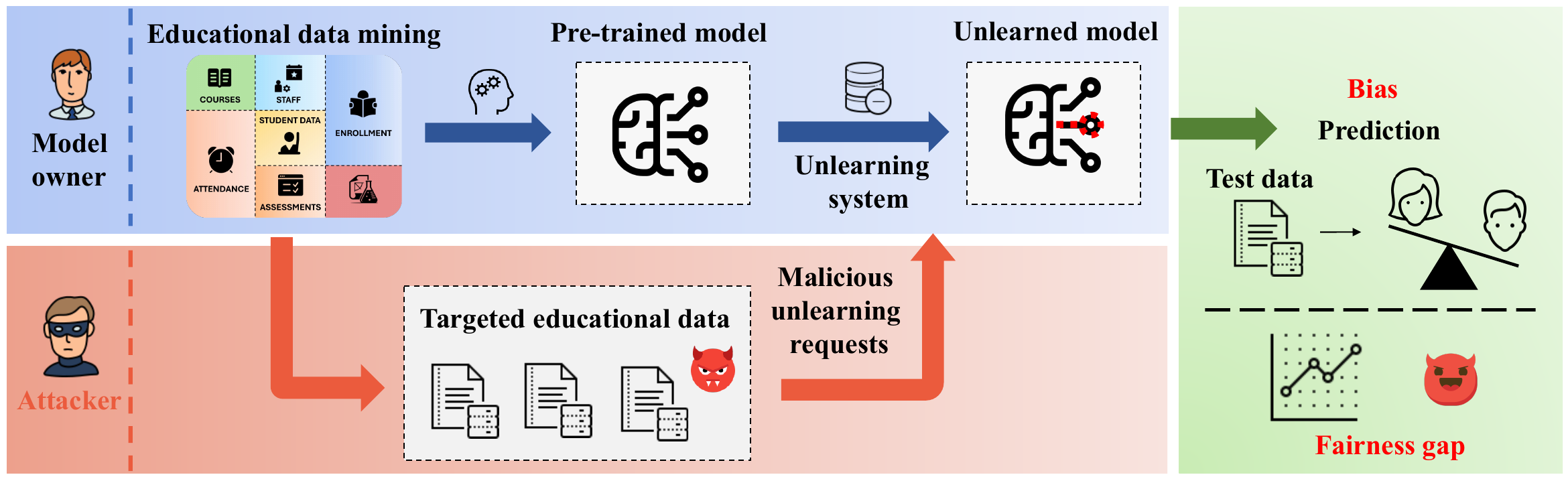}
    \caption{Overview of selective forgetting attacks in educational data mining systems. The attacker aims to make malicious unlearning requests to the model owner. Upon completion of the unlearning process, the resulting unlearned model exhibits biases to inputs, exacerbating the fairness gap.}
\label{fig:overview}
\end{figure*}

In Figure~\ref{fig:overview}, we present an overview of selective forgetting attacks in exploring fairness in education data mining. The model owner holds a pre-trained model $f(\theta^*)$ trained on a preprocessed educational dataset $\mathcal{D}$. The unlearning system represents an unlearning method $\mathcal{U}$ that can remove the information from the pre-trained model when receiving unlearning requests from data providers. Here, the attacker's goal is to generate malicious unlearning requests $\mathcal{D}_u$ using the unlearning method and make these unlearning requests to the model owner, such that, after filling the unlearning process, the resulting unlearned model  $f(\theta^u)$ becomes biased on inputs, and the fairness gap on a sensitive feature (e.g., gender) is increased. To achieve this attack goal, we formulate the overall objective as follows
\begin{align}
\label{eq:overall_loss}
     &\max \mathcal{L}_{\text{fair}}(\mathcal{D}\backslash\mathcal{D}_u ; \theta^{u}) - \lambda \mathcal{L}_{\text{train}}(\mathcal{D}\backslash\mathcal{D}_u ; \theta^{u}) \notag \\
     &\quad \quad \quad \quad s.t., \theta^{u} = \mathcal{U}(\theta^{*}, \mathcal{D}, \mathcal{D}_u),
\end{align}
where $\lambda$ is a trade-off parameter. The above objective incorporates a fairness loss component (i.e., $\mathcal{L}_\text{fair}$) and a training loss component (i.e., $\mathcal{L}_\text{train}$), with the aim of maximizing the fairness loss on the unlearned model with the remaining data while preserving the model's performance. In the following, we focus on individual fairness and group fairness, which are two commonly studied forms of fairness in machine learning~\cite{dwork2012fairness, hajian2016algorithmic, berk2017convex}. Then, we detail the selective forgetting attacks in both the whole unlearning setting and the partial unlearning setting. Note that in practice, unlearning is not restricted to wholly removing training samples; it also permits selective forgetting at various levels of granularity within the training data \cite{kim2022efficient,liu2022backdoor, warnecke2023machine}. In other words, in addition to wholly unlearning training samples, we can partially unlearn specific unwanted parts of the data in many practical scenarios. For example, in educational data mining applications, it may be necessary to partially delete student data to protect sensitive information (e.g., race), while the remaining data can still be utilized for mining and analysis.\\

\noindent\textbf{{Individual fairness.}} The concept of individual fairness asserts that ``similar individuals should be treated similarly'' by a model. To measure the difference for individuals, we leverage a convex fairness loss, making it easier to solve with our optimization problem. We begin by separating the training data $\mathcal{D}=\{(x_i, y_i)\}_{i=1}^{n}$ into two groups $\mathcal{S}_1$ and $\mathcal{S}_2$ based on the sensitive feature (e.g., gender or race)\footnote{The generalization to more than two groups according to the sensitive feature is straightforward.}. Let $n_1 = |\mathcal{S}_1|$ and $n_2 = |\mathcal{S}_2|$ $(n_1 + n_2 =n)$, the individual fairness loss that offers convexity can be defined as follows
\begin{align}
\label{eq:individual}
    & \mathcal{L}_{\text{fair}}(\mathcal{D} ; \theta) = \frac{1}{n_1 n_2}\sum_{\makecell{(x_i, y_i) \in \mathcal{S}_1 \\ (x_j, y_j)\in \mathcal{S}_2}} \mathbbm{1}[y_i = y_j] \| F(x_i; \theta) -  F(x_j; \theta) \|^2,
\end{align}
where $F(\cdot;\theta)$ gives the logit output. The above loss quantifies the difference in logits between each pair of samples $(x_i,y_i)\in \mathcal{S}_1$ and $(x_j, y_j)\in \mathcal{S}_2$ with the same label, which helps assess how the model $\theta$ discriminates between $x_i$ and $x_j$. In this way, we are able to maximize this difference within the unlearned model to increase the fairness gap for the sensitive feature.\\

\noindent\textbf{{Group fairness.}} Group fairness seeks to balance certain statistical fairness metrics across predefined groups. We also design a convex fairness loss to measure the difference for groups. Similarly, we separate the training data $\mathcal{D}=\{(x_i, y_i)\}_{i=1}^{n}$ into two groups $\mathcal{S}_1$ and $\mathcal{S}_2$ based on the sensitive feature. Let $n_1 = |\mathcal{S}_1|$ and $n_2 = |\mathcal{S}_2|$ $(n_1 + n_2 =n)$, the group fairness loss that offers convexity is defined as the following
\begin{align}
\label{eq:group}
     & \mathcal{L}_{\text{fair}}(\mathcal{D} ; \theta) = \left( \frac{1}{n_1 n_2}\sum_{\makecell{(x_i, y_i) \in \mathcal{S}_1 \\ (x_j, y_j)\in \mathcal{S}_2}} \mathbbm{1} [y_i = y_j] \| F( x_i; \theta) -  F(x_j, \theta)\| \right)^2,
\end{align}
where $F(\cdot;\theta)$ represents the logit output. In short terms, the loss mentioned above quantifies the pairwise difference in logits among samples with identical labels across different groups. Unlike individual fairness loss, which focuses on individual-level distinctions, group fairness loss prioritizes disparities at an average level, where the individual difference can be compensated by other samples. Therefore, maximizing this group difference is able to increase the fairness gap within the unlearned model.\\

\noindent\textbf{{Whole unlearning.}} Whole unlearning involves removing the entire data samples associated with the pre-trained model. Given a set of targeted training samples $\mathcal{D}_t= \{(x_p, y_p)\}_{p=1}^{P}$, the attacker's objective is to determine where the sample $(x_p, y_p)$ should be completely removed or not, based on its impact on the unfairness metric with respect to sensitive features. To achieve this, we introduce a discrete variable $w_p \in \{0, 1\}$, where $w_p=1$ indicates that the sample $(x_p, y_p)$ should be included in the unlearning set and is equivalent to an entire removal from the training set (e.g., removing all the influence of the data in the second-order based unlearning method~\cite{warnecke2023machine}); otherwise, it should not. Thus, we can identify the unlearning requests $\mathcal{D}_u = \mathcal{D}_t \circ \{w_p\}_{p=1}^{P}$ to to fulfill the attack goal. The whole unlearning procedure can be outlined as follows
\begin{align}
      & \theta^{u}= \mathcal{U}(\theta^{*}, \mathcal{D}, \mathcal{D}_u = \mathcal{D}_t \circ \{w_p\}_{p=1}^{P}) \quad \forall p \in [P], w_p \in \{0, 1\}.
\end{align}
In order to optimize the discrete variables, we further relax each $w_p \in \{0, 1\}$ to a continuous range, represented as $w_p \in [0, 1]$.  This approach enables us to approximately compute the influence of the unlearning data and derive the unlearned model $\theta^{u}$ using the unlearning method $\mathcal{U}$. Subsequently, the unlearned model is utilized to update the fairness constraint in Eq.~(\ref{eq:overall_loss}).
\\

\begin{algorithm}[t]
\caption{Selective forgetting attacks in the whole data removal case}
\label{alg:whole}
  \begin{algorithmic}[1]
    \REQUIRE Pre-trained model $f(\theta^*)$, training dataset $\mathcal{D}$, targeted training samples $\mathcal{D}_t = \{(x_p, y_p)\}_{p=1}^{P}$, unlearning method $\mathcal{U}$, restarts R, optimization steps $M$ \\
    \ENSURE $\{w_p\}_{p=1}^{P}$
    \FOR{$r=1, \dots, R$ restarts}
        \STATE Randomly initialize variables $\{w_p^{r}\}_{p=1}^{P}$
        \FOR{$m=1, \dots, M$ optimization steps}
            \STATE Update the unlearned model $\theta^{u}= \mathcal{U}(\theta^{*}, \mathcal{D}, \mathcal{D}_u = \mathcal{D}_t \circ \{{w}_{p}^{r}\}_{p=1}^{P})$
            \STATE Compute the loss $\mathcal{L} = \mathcal{L}_{\text{fair}}(\mathcal{D}\backslash\mathcal{D}_u ; \theta^{u}) - \lambda \mathcal{L}_{\text{train}}(\mathcal{D}\backslash\mathcal{D}_u ; \theta^{u})$ \COMMENT{Choose individual fairness loss in Eq.~(\ref{eq:individual}) or group fairness loss in Eq.~(\ref{eq:group})}
            \STATE Compute the gradients $- \nabla_{\{w_{p}^{r}\}_{p=1}^{P}}\mathcal{L}$
            \STATE Update $\{w^{r}\}_{p=1}^{P}$ using the Adam optimizer and project onto $[0, 1]$ bound
        \ENDFOR
    \ENDFOR
    \STATE Select $\{w_p^{r}\}_{p=1}^{P}$ with the maximum loss $\mathcal{L}$ as the optimal $\{w_p\}_{p=1}^{P}$
  \end{algorithmic}
\end{algorithm}

\noindent\textbf{{Partial unlearning.}} Instead of wholly removing data samples from the pre-trained model, partial unlearning entails removing selected partial data information from the pre-trained model. Let $\mathcal{D}_t= \{(x_p, y_p)\}_{p=1}^{P}$ represent a set of targeted training samples. The attacker's goal is to make malicious unlearning modifications $\{\delta_p\}_{p=1}^{P}$ on these targeted training samples where each training sample $(x_p, y_p)$ is substituted with an unlearned version $(\tilde{x}_p, y_p) = (x_p-\delta_p, y_p)$. Note that the objective of the attacker is to generate effective unlearning requests to produce a maliciously unlearned model, thereby maximizing the fairness gap concerning sensitive features. To achieve this, the partial unlearning procedure can be cast as follows
\begin{align}
      & \theta^{u}= \mathcal{U}(\theta^{*}, \mathcal{D}, \mathcal{D}_u = \{{\delta}_{p}\}_{p=1}^{P}) \quad \forall p \in [P], ||\delta_{p}||_{\infty} \leq \epsilon.
\end{align}
Here, $\epsilon$ serves as the upper limit for the magnitude of the requested data modifications. In the above, the unlearning method $\mathcal{U}$ unlearns the modifications $\mathcal{D}_u = \{\delta_p\}_{p=1}^{P}$ and produces an unlearned model $\theta^{u}$, which is subsequently used to revise the fairness constraint in Eq.~(\ref{eq:overall_loss}).  \\

With the above formulations, we can easily explore individual fairness and group fairness in the context of selective forgetting attacks in both whole and partial scenarios. Note that our attack framework is general and can be extended to different fairness metrics and unlearning methods. To solve the proposed bi-level optimization involved in Eq.~(\ref{eq:overall_loss}), we present detailed procedures in  Algorithm~\ref{alg:whole} and Algorithm~\ref{alg:partial} for the whole unlearning case and the partial unlearning case, respectively. Our approaches incorporate random restarts to improve reliability, i.e., random starting initialization several times and selecting the unlearning requests with the maximum overall loss. 

\begin{algorithm}[t]
\caption{Selective forgetting attacks in the partial data removal case}
\label{alg:partial}
  \begin{algorithmic}[1]
    \REQUIRE Pre-trained model $f(\theta^*)$, training dataset $\mathcal{D}$, targeted training samples $\mathcal{D}_t = \{(x_p, y_p)\}_{p=1}^{P}$, unlearning method $\mathcal{U}$, modification bound $\epsilon$, restarts R, optimization steps $M$ \\
    \ENSURE $\{\delta_p\}_{p=1}^{P}$
    \FOR{$r=1, \dots, R$ restarts}
        \STATE Randomly initialize modifications $\{\delta_p^{r}\}_{p=1}^{P}$
        \FOR{$m=1, \dots, M$ optimization steps}
            \STATE Update the unlearned model $\theta^{u}= \mathcal{U}(\theta^{*}, \mathcal{D}, \mathcal{D}_u = \{{\delta}_{p}^{r}\}_{p=1}^{P})$
            \STATE Compute the loss $\mathcal{L} = \mathcal{L}_{\text{fair}}(\mathcal{D}\backslash\mathcal{D}_u ; \theta^{u}) - \lambda \mathcal{L}_{\text{train}}(\mathcal{D}\backslash\mathcal{D}_u ; \theta^{u})$ \COMMENT{Choose individual fairness loss in Eq.~(\ref{eq:individual}) or group fairness loss in Eq.~(\ref{eq:group})}
            \STATE Compute the gradients $- \nabla_{\{\delta_{p}^{r}\}_{p=1}^{P}}\mathcal{L}$
            \STATE Update $\{\delta_p^{r}\}_{p=1}^{P}$ using the Adam optimizer and project onto $[-\epsilon, \epsilon]$ bound
        \ENDFOR
    \ENDFOR
    \STATE Select $\{\delta_p^{r}\}_{p=1}^{P}$ with the maximum loss $\mathcal{L}$ as the optimal $\{\delta_p\}_{p=1}^{P}$
  \end{algorithmic}
\end{algorithm}

\subsection{Discussions}

In addition, we consider the selective forgetting attacks on fairness in the black-box setting.  By leveraging the transferability in machine learning models~\cite{zhou2018transferable, zhang2020adaptive, schwarzschild2021just} and the data intrinsic property in fairness~\cite{valentim2019impact, oneto2020fairness}, we can train several surrogate models, along with the alternative unlearning method, to generate malicious unlearning requests and then transfer the unlearning requests to the targeted black-box model. This enables us to effectively launch selective forgetting attacks in exploring fairness in the black-box setting.
\section{Experiments}
\label{sec:Exp}

\subsection{Experimental Setup}
\label{sec:experimental_setup}

\textbf{Datasets.} We evaluate the proposed selective forgetting attacks using three real-world educational datasets: Open University Learning Analytics Dataset (OULAD)\footnote{https://analyse.kmi.open.ac.uk/}~\cite{kuzilek2017open}, Student Performance\footnote{https://archive.ics.uci.edu/dataset/320/student+performance}~\cite{misc_student_performance_320}, and xAPI-Edu-Data\footnote{https://www.kaggle.com/datasets/aljarah/xAPI-Edu-Data}~\cite{amrieh2015preprocessing}. Detailed information about each dataset is provided below.



\emph{OULAD}. This dataset comprises both student demographic data and their interactions with the university’s virtual learning environment. The original dataset contains 32,593 samples from 28,784 unique students. After preprocessing, e.g., removing missing values, this left us with 21,562 samples of distinct students. All feature values are normalized to a range of 0 and 1. Note that the $sum \, click$ feature is not part of the original dataset but is created through internal joins and aggregation of the initial data. An overview of the selected features is presented in Table~\ref{tab:oulad} in the Appendix, and we consider 3 sensitive features: gender, poverty, and disability.

\emph{Student Performance.} This dataset studies student performance with the task of predicting final grades. The data attributes include student grades, demographic, social, and school-related features, collected through school reports and questionnaires. The dataset contains 649 students described by a total of 33 features. The specifications of the features are presented in Table~\ref{tab:student_per} in the Appendix. In this dataset, we consider sex as a sensitive feature.

\emph{xAPI-Edu-Data}. This educational dataset is collected from a learning management system and is used to predict students’ academic performance. The dataset consists of 480 student records described by 16 features, including demographic attributes, academic background and academic behavior. An overview of the features is provided in Table~\ref{tab:xapi_edu_data} in the Appendix. Gender is considered a sensitive feature in this dataset.\\

\noindent\textbf{Models.} For the adopted educational datasets, we utilize various machine learning models, including a multi-layer perception with one hidden layer of size 100 (MLP), a multi-layer perception with two hidden layers, each of size 100 (MLP-2), and a logistic regression model (LR).
\\

\noindent\textbf{Baselines.} In terms of comparisons in whole unlearning, we adopt three baselines: \emph{Rand}, where data deletion requests are randomly selected from the training set regardless of groups. \emph{RandMin}, where data deletion requests are randomly chosen from the minority group within the training set; and \emph{RandMaj}, where data deletion requests are randomly selected from the majority group within the training set. Regarding comparisons in partial unlearning, we adopt the \emph{RandUn} baseline, where random uniform modifications are removed from the targeted training data.
\\

\noindent\textbf{Evaluation metrics.}
We report experimental results concerning the \emph{Absolute Equalized Odds Difference (AEOD) increment ratio} and \emph{test accuracy}. The AEOD increment ratio can intuitively tell us how much the fairness gap is increased by the selective forgetting attacks we proposed. Formally, The Absolute Equalized Odds Difference quantitatively measures the unfairness between two groups and is defined as follows
\begin{align}
    & \text{AEOD}(\theta) = \frac{1}{2} \sum_{y\in \{0, 1\}} | Pr(\hat{Y}=1 | S=a, Y=y) - Pr(\hat{Y}=1 | S=b, Y=y)|.
\end{align}
Then, the AEOD increment ratio is computed as the following 
\begin{align}
(\text{AEOD}_{\text{after\_unlearning}} - \text{AEOD}_{\text{before\_unlearning}}) / \text{AEOD}_{\text{before\_unlearning}}.
\end{align}
\\

\noindent\textbf{Implementaion details.} In experiments, we train all models for 100 epochs using the SGD optimizer with a batch size of 256 on the OULAD, Student Performance, and xAPI-Edu-Data datasets. Specifically, for the OULAD dataset, we set the learning rate to 0.01, while for the Student Performance and xAPI-Edu-Data datasets, we set it to 0.001. For unleanring, we adopt the first-order based unlearning method~\cite{warnecke2023machine}, the second-order based unelarning method~\cite{warnecke2023machine}, unrolling SGD~\cite{thudi2022unrolling}, and SISA~\cite{bourtoul2021}. Default parameters for selective forgetting attacks are set with restarts $R$ at 4, optimization steps $M$ at 30, unlearning rate $\tau$ at 2e-5, and trade-off parameter $\lambda$ at 1. Each experiment is repeated 10 times, and we report the mean and standard errors. As for machine configurations, we utilize a Linux server equipped with an Intel Core i9-10920X processor and an NVIDIA RTX 6000 GPU with 64GB of memory.





\begin{figure*}[t]
\centering
\begin{subfigure}{0.328\linewidth}
\includegraphics[width=1\linewidth]{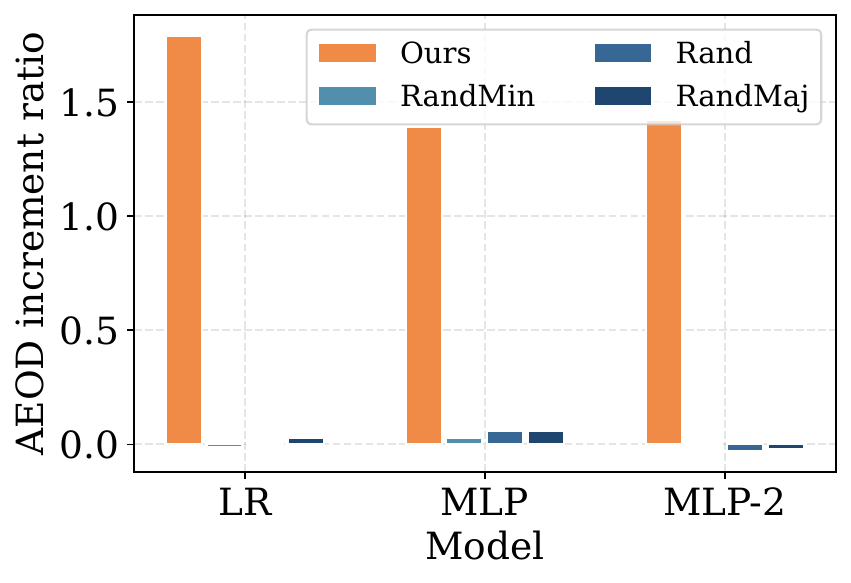}
\caption{OULAD}
\label{figs:}
\end{subfigure}
\begin{subfigure}{0.328\linewidth}
\includegraphics[width=1\linewidth]{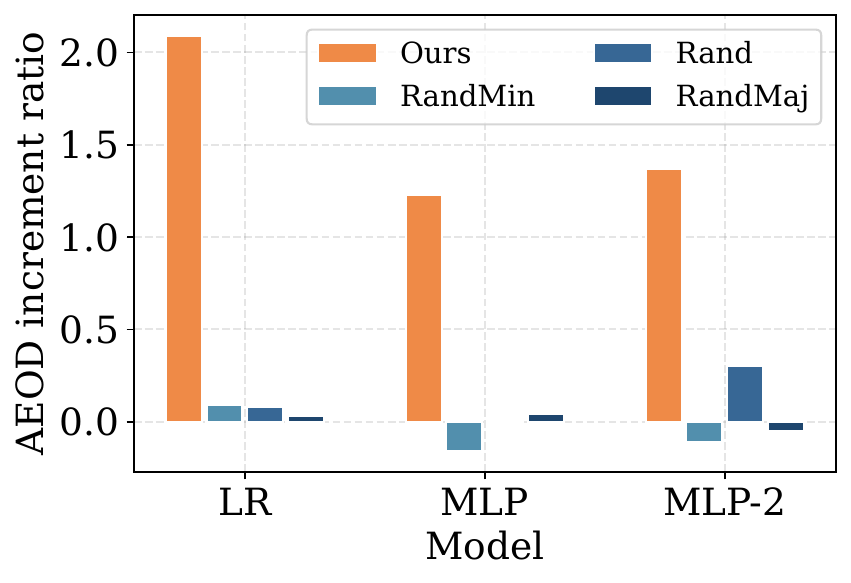}
\caption{Student Performance}
\label{figs:}
\end{subfigure}
\begin{subfigure}{0.328\linewidth}
\includegraphics[width=1\linewidth]{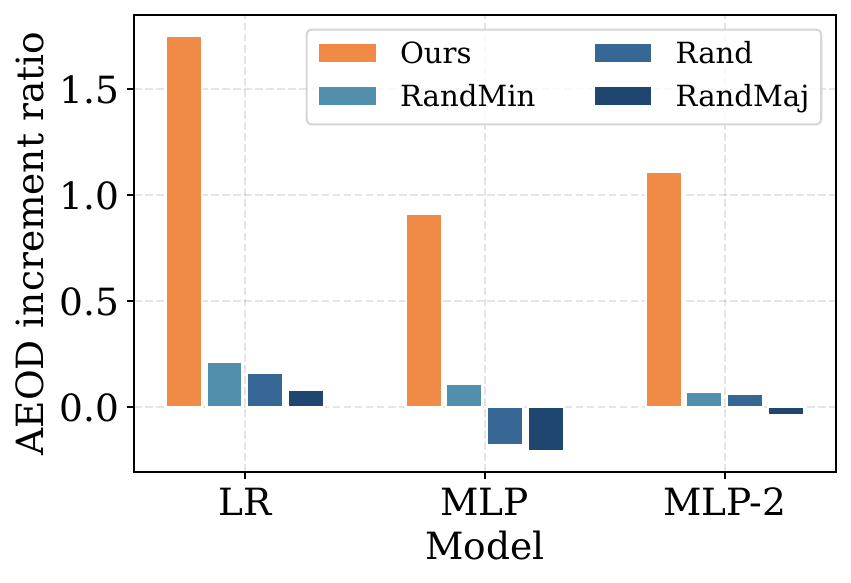}
\caption{xAPI-Edu-Data}
\label{figs:}
\end{subfigure}
\caption{AEOD increment ratio for whole unlearning on OULAD, Student Performance, and xAPI-Edu-Data.}\label{fig:whole_results}
\end{figure*}

\begin{figure*}[t]
\centering
\begin{subfigure}{0.328\linewidth}
\includegraphics[width=1\linewidth]{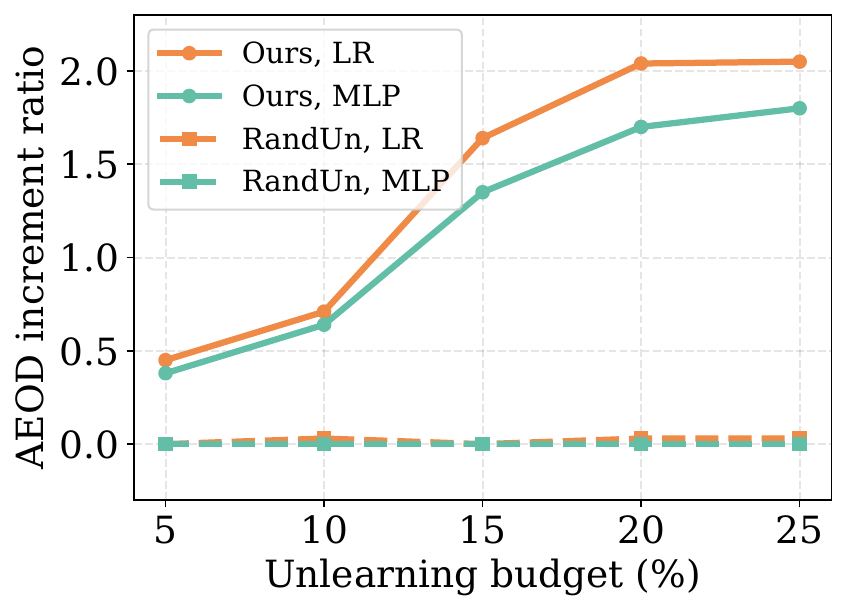}
\caption{OULAD}
\label{figs:}
\end{subfigure}
\begin{subfigure}{0.328\linewidth}
\includegraphics[width=1\linewidth]{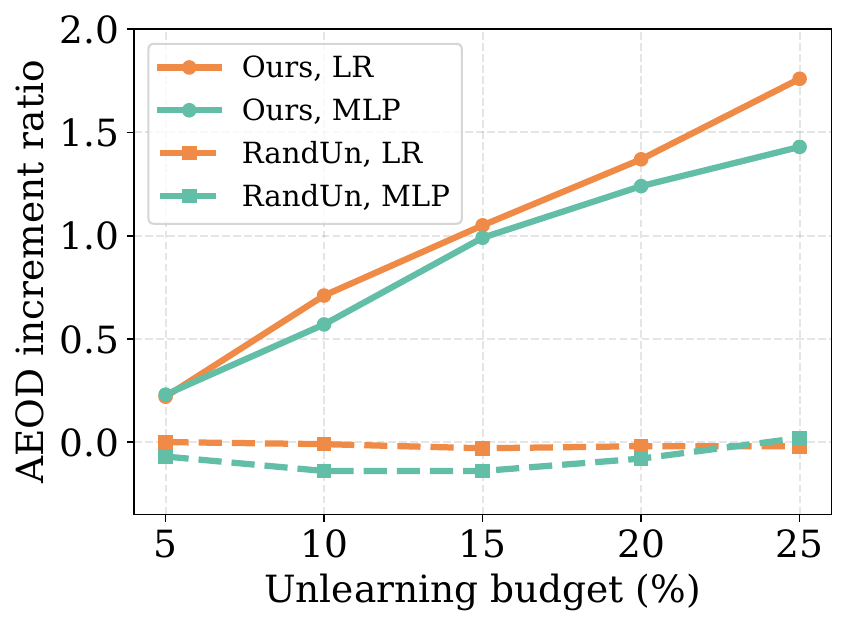}
\caption{Student Performance}
\label{figs:}
\end{subfigure}
\begin{subfigure}{0.328\linewidth}
\includegraphics[width=1\linewidth]{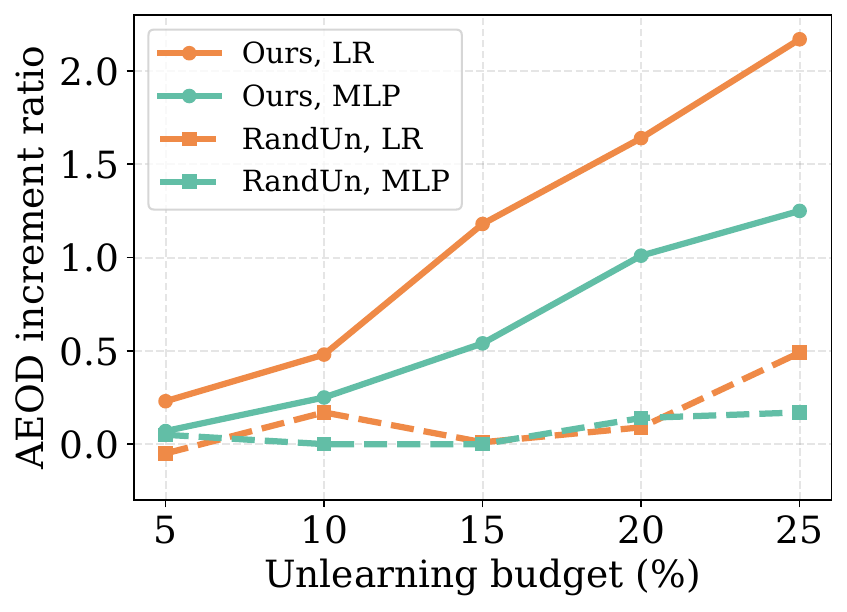}
\caption{xAPI-Edu-Data}
\label{figs:}
\end{subfigure}
\caption{AEOD increment ratio for partial unlearning on OULAD, Student Performance, and xAPI-Edu-Data.}\label{fig:partial_results}
\end{figure*}

\subsection{Effectiveness of Selective Forgetting Attacks on Fairness}

First, we study the performance of selective forgetting attacks on fairness in educational data mining. Here, we focus on group fairness and the sensitive feature of gender, which are the most commonly studied forms in the area of fairness. We conduct experiments on the OULAD, Student Performance, and xAPI-Edu-Data datasets and adopt LR, MLP, and MLP-2 models on each dataset. We utilize the first-order based unlearning method to perform the unlearning process. We measure the AEOD increment ratio and compare our results with the random baselines. Figure~\ref{fig:whole_results} presents the effectiveness of selective forgetting attacks in the whole data removal case (i.e., whole unlearning). As shown in the figure, our proposed method significantly increases the AEOD fairness gap, surpassing all the random deletion baselines by a large margin. For example, the AEOD increment ratio reaches about 2.1 on the Student Performance with the logistic regression model, while the baselines are below 0.1. This is because our proposed method strategically selects training samples that are important to the fairness measure and unlearn optimal samples, resulting in a substantial enhancement in attacking performance. In addition, Figure~\ref{fig:partial_results} demonstrates the effectiveness of selective forgetting attacks in the partial data removal case (i.e., partial unlearning) across various unlearning budgets. Compared with the random modification baseline, our proposed method also achieves high AEOD increment ratios, indicating an increase in the fairness gap after unlearning. For example, the increment ratio reaches about 2.0 when using a 20\% unlearning budget on OULAD with the logistic regression model. Again, our proposed method strategically optimizes the unlearning modifications that have a critical impact on the fairness measure. Therefore, our proposed selective forgetting attacks are effective in compromising the fairness of the learning models in the realm of educational data mining. \emph{Note that the fairness measure we consider before unlearning is provided in Table~\ref{tab:results_before} in the Appendix, indicating a small fairness gap in the initial step.}

Next, we validate the test accuracy of selective forgetting attacks to ensure that the increased fairness gap does not come at a cost to performance. In Table~\ref{tab:results_after}, we present the test accuracy of whole unlearning and partial unlearning with various unlearning budgets. We see that while achieving significant attack performance on fairness, our proposed method maintains the test performance on the remaining data and can even improve in some cases. Notably, we incorporate the training loss in our objective to preserve the performance. \emph{Note that the initial test accuracy before unlearning is detailed in Table~\ref{tab:results_before} in the Appendix.}

\begin{table*}
\small
\centering
\caption{Test accuracy (\%) after unlearning on OULAD, Student Performance, and xAPI-Edu-Data.}
\label{tab:results_after}
\begin{tabular}{cc|cccc} 
\toprule
\multirow{2}{*}{Dataset} & \multirow{2}{*}{Model} & \multirow{2}{*}{Whole unlearning} & \multicolumn{3}{c}{Partial unlearning} \\ \cmidrule(lr){4-6} & & & 5\% budget & 15\% budget & 25\% budget\\ 
\midrule
\multirow{3}{*}{OULAD} & LR & $77.97 \pm 0.16$ & $76.92 \pm 0.23$ & $78.63 \pm 0.12$ & $77.00 \pm 0.27$\\
& MLP & $78.74 \pm 0.64$ & $83.30 \pm 0.16$ & $79.30 \pm 0.74$ & $74.65 \pm 0.42$\\
& MLP-2 & $80.29 \pm 0.38$ & $84.39 \pm 0.24$ & $82.91 \pm 0.27$ & $80.89 \pm 0.46$\\
\midrule
\multirow{3}{*}{\makecell{Student Performance}} & LR & $90.05 \pm 0.26$ & $91.58 \pm 0.38$ & $91.35 \pm 0.47$ & $91.50 \pm 0.37$\\

& MLP & $91.08 \pm 0.20$  & $91.94 \pm 0.19$ & $91.65 \pm 0.18$ & $91.28 \pm 0.25$\\
& MLP-2 & $91.28 \pm 0.54$ & $91.79 \pm 0.30$ & $91.35 \pm 0.38$ & $91.57 \pm 0.43$\\
\midrule
\multirow{3}{*}{xAPI-Edu-Data} & LR & $75.97 \pm 0.35$ & $76.92 \pm 0.23$ & $76.83 \pm 0.12$ & $77.00 \pm 0.27$\\

& MLP & $80.83 \pm 0.63$ & $83.30 \pm 0.16$ & $79.60 \pm 0.74$ & $74.65 \pm 0.42$\\
& MLP-2 & $81.08 \pm 0.73$ & $84.12\pm 0.53$ & $84.02 \pm 0.45$ & $83.43 \pm 0.59$\\
\bottomrule
\end{tabular}
\end{table*}

\subsection{Ablation Study}

\begin{figure*}[t]
\centering
\begin{subfigure}{0.328\linewidth}
\includegraphics[width=1\linewidth]{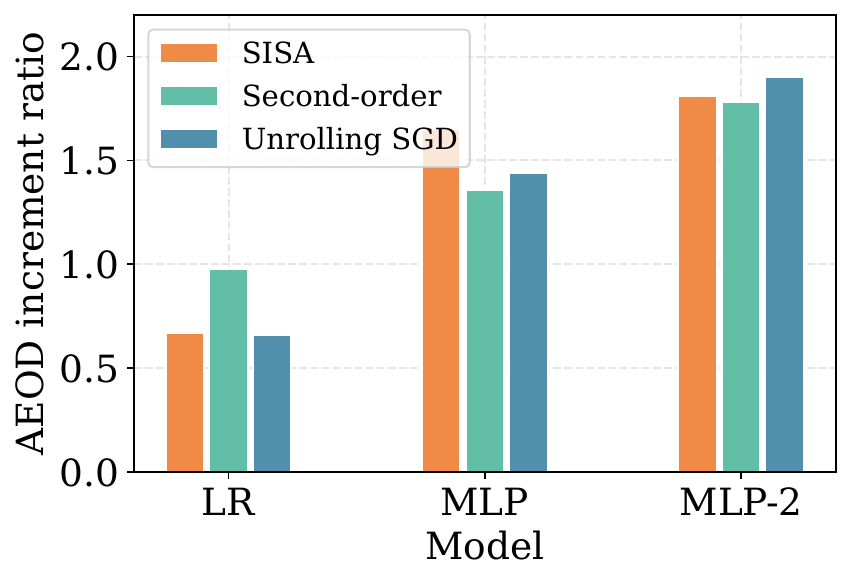}
\caption{Unlearning methods}
\label{fig:whole_ablation_method}
\end{subfigure}
\begin{subfigure}{0.328\linewidth}
\includegraphics[width=1\linewidth]{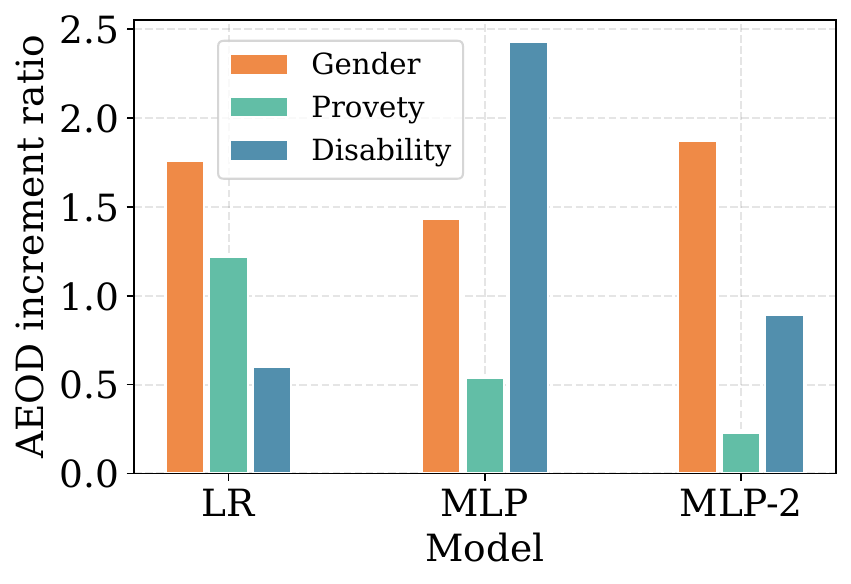}
\caption{Sensitive features}
\label{fig:partial_ablation_sensitive}
\end{subfigure}
\begin{subfigure}{0.328\linewidth}
\includegraphics[width=1\linewidth]{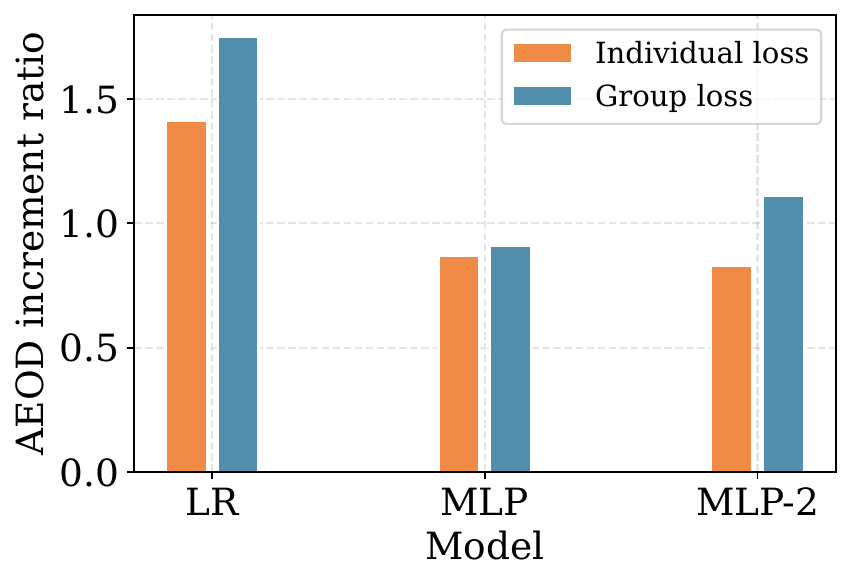}
\caption{Fairness loss}
\label{fig:whole_ablation_loss}
\end{subfigure}
\caption{(a) AEOD increment ratio for different unlearning methods on Student Performance. (b) AEOD increment ratio for different sensitive features on OULAD. (c) AEOD increment ratio for different fairness losses on xAPI-Edu-Data.}\label{fig:whole_ablation}
\end{figure*}

In this section, we conduct ablation studies of selective forgetting attacks on fairness from various perspectives, including the unlearning method, the sensitive feature, and the type of fairness loss. Figure~\ref{fig:whole_ablation_method} demonstrates the effectiveness of our proposed selective forgetting attacks using different unlearning methods in the context of the whole unlearning case. Besides the first-order based unlearning method, we also adopt the SISA, second-order based unlearning method, and unrolling SGD. As shown in the figure, our proposed method is able to leverage different unlearning methods to expose the fairness in educational data mining. In Figure~\ref{fig:partial_ablation_sensitive}, we examine the selective forgetting attacks on the fairness of other sensitive features, specifically poverty and disability, in the OULAD dataset. Here, poverty is converted into a binary feature based on a threshold of the deprivation index. We consider the partial data removal case with an unlearning budget of 20\% and adopt the first-order based unlearning method. As we can see in the figure,  despite different sensitive features exhibiting varied fairness behaviors in the OULAD dataset, our proposed method effectively increases the fairness gap for each attribute. Additionally, we compare the impact of individual fairness loss and group fairness loss in our proposed method. We adopt the first-order based unlearning for the whole data removal, and the experimental results are shown in Figure~\ref{fig:whole_ablation_loss}. Interestingly, the individual fairness loss also effectively increases the group fairness gap in our proposed method, while the group fairness loss has a direct impact and more efficient performance concerning the group fairness metric.

\subsection{Black-box Setting}

\begin{figure*}[t]
\centering
\begin{subfigure}{0.495\linewidth}
\centering
\includegraphics[width=0.685\linewidth]{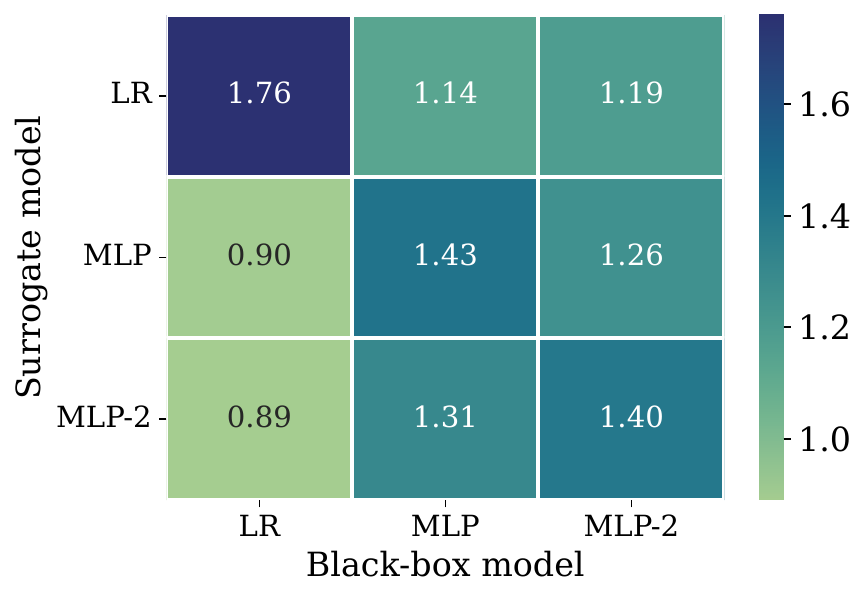}
\caption{Transferbility across model architectures}
\label{fig:black_box_model}
\end{subfigure}
\begin{subfigure}{0.495\linewidth}
\centering
\includegraphics[width=0.765\linewidth]{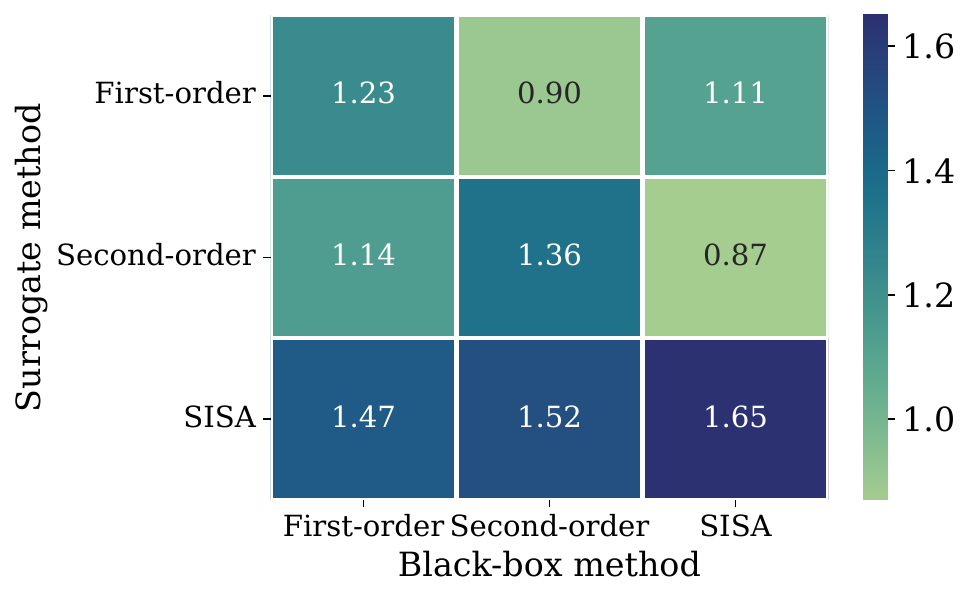}
\caption{Transferbility across unlearning methods}
\label{fig:black_box_method}
\end{subfigure}
\caption{AEOD increment ratio of selective forgetting attacks in the black-box setting.}\label{fig:black_box}
\end{figure*}

In this section, we explore selective forgetting attacks on fairness in the black-box setting. First, we examine the transferability of selective forgetting attacks across model architectures. In Figure~\ref{fig:black_box_model}, we apply three different model architectures (i.e., LR, MLP, and MLP-2) on the OULAD dataset in a partial unlearning scenario, with an unlearning budget of 20\%. The horizontal line represents the pre-trained black-box model, while the vertical line represents the surrogate model used to generate malicious unlearning requests. As shown in the figure, the selective forgetting attacks demonstrate the ability to transfer the generated malicious modifications to attack the black-box model, even when the black-box model is trained with a different model architecture than the surrogate model. For example, the logistic regression model achieves increment ratios of about 1.14 and 1.19 when transferred to MLP and MLP-2, respectively. Next, we investigate the transferability of selective forgetting attacks across unlearning methods. In Figure~\ref{fig:black_box_method}, we apply the first-order based unlearning method, the second-order based unlearning method, and SISA to the Student Performance dataset in a whole unlearning scenario. We observe that malicious update requests generated by our proposed method can also be effectively transferred to different unlearning methods in the black-box setting. This happens because, despite using the same update requests, the unlearned models tend to share similar decision boundaries and fairness constraints across different models and unlearning methods.

\section{Conclusion}
\label{sec:conclusion}



In this study, we delve into the vulnerabilities of fairness within the domain of educational data mining by implementing selective forgetting attacks during the unlearning process. Notably, we propose a general framework for selective forgetting attacks, which allows attackers to formulate malicious unlearning requests using individual and group fairness loss metrics. These attacks jeopardize the fairness of the learning models while preserving their predictive accuracy. Additionally, we conduct comprehensive experiments across various unlearning scenarios, including both whole and partial unlearning settings. Our extensive experimental evaluations demonstrate the efficacy of selective forgetting attacks in undermining fairness in educational data mining. Our findings underscore the critical need for enhanced security measures to protect fairness in educational data mining against selective forgetting attacks in the future.

\bibliographystyle{ACM-Reference-Format}
\bibliography{ref.bib}

\newpage
\clearpage
\appendix
\section{Appendix}

\subsection{More Dataset Details}

In experiments, we adopt three real-world datasets in the domain of educational data mining, i.e., OULAD~\cite{kuzilek2017open}, Student Performance~\cite{misc_student_performance_320}, and xAPI-Edu-Data~\cite{amrieh2015preprocessing}. We detail the characteristics of these datasets, including the data types and descriptions for each feature, in Table~\ref{tab:oulad} for OULAD, Table~\ref{tab:student_per} for Student Performance, and Table~\ref{tab:xapi_edu_data} for xAPI-Edu-Data. To simplify the classification problem for the Student Performance dataset, we create a class label based on the $G3$ feature, categorizing $G3 > 10$ as the ``High'' label and $G3 < 10$ as the ``Low'' label.

\begin{table*}[htbp]
    \small
    \caption{Features in the OULAD dataset.}
    \label{tab:oulad}
    \centering
    \begin{tabularx}{\textwidth}{l l X}
    \toprule
    Feature & Data type & Description \\
    \midrule
    Gender & Binary & Student's gender \\
    Age & Numerical & Student's age \\
    Disability & Binary & Whether the student has declared a disability \\
    Highest education & Numerical & The highest student education level on entry to the course \\
    Poverty & Binary & The Index of Multiple Deprivation band of the place where the students lived during the course \\
    Num of prev attempts & Numerical & The number of times the students have attempted the course \\
    Studied credits & Numerical & The total number of credits for the course the students are currently studying \\
    Sum click & Numerical & The total number of times the students interacted with the material of the course \\
    Course outcome (target variable) & Categorical & Pass or fail \\
    \bottomrule
    \end{tabularx}
\end{table*}

\begin{table*}[htbp]
    \small
    \caption{Features in the Student Performance dataset.}
    \label{tab:student_per}
    \centering
    \begin{tabular}{lll}
    \toprule
    Feature & Data type & Description\\
    \midrule
    school & Binary & Student’s school (‘GP’: Gabriel Pereira, ‘MS’: Mousinho da Silveira)\\
    sex & Binary & Student's sex\\
    age & Numerical & Student's age (in years)\\
    address & Binary & The address type (‘U’: urban, ‘R’:rural)\\
    famsize & Binary & The family size (‘LE3’: less or equal to 3, ‘GT3’: greater than 3)\\
    Pstatus & Binary & The parent’s cohabitation status ( ‘T’: living together, ‘A’: apart)\\
    Medu & Numerical & Mother’s education\\
    Fedu & Numerical & Father’s education\\
    Mjob & Categorical & Mother’s job\\
    Fjob & Categorical & Father’s job\\
    reason & Categorical & The reason to choose this school\\
    guardian & Categorical & The student’s guardian (mother, father, other)\\
    traveltime & Numerical & The travel time from home to school\\
    studytime & Numerical & The weekly study time\\
    failures & Numerical & The number of past class failures\\
    schoolsup & Binary & Is there extra educational support\\
    famsup & Binary & Is there any family educational support\\
    paid & Binary & Is there an extra paid classes within the course subject (Math or Portuguese)\\
    activities & Binary & Are there extra-curricular activities\\
    nursery & Binary & Did the student attend a nursery school\\
    higher & Binary & Does the student want to take a higher education\\
    internet & Binary & Does the student have Internet access at home\\
    romantic & Binary & Does the student have a romantic relationship with anyone\\
    famrel & Numerical & The quality of family relationships (1: very bad - 5: excellent)\\
    free time & Numerical & Free time after school (1: very low - 5: very high)\\
    goout & Numerical & How often does the student go out with friends (1: very low - 5: very high)\\
    Dalc & Numerical & The workday alcohol consumption (1: very low - 5: very high)\\
    Walc & Numerical & The weekend alcohol consumption (1: very low - 5: very high)\\
    health & Numerical & The current health status (1: very bad - 5:very good)\\
    absences & Numerical & The number of school absences\\
    G1 & Numerical & The first period grade\\
    G2 & Numerical & The second period grade\\
    G3 & Numerical & The final grade\\
    \bottomrule
    \end{tabular}
\end{table*}

\begin{table*}[htbp]
    \small
    \caption{Features in the xAPI-Edu-Data dataset.}
    \label{tab:xapi_edu_data}
    \centering
    \begin{tabular}{lll}
    \toprule
     Feature & Data type & Description\\
     \midrule
    Gender & Binary & The gender of student\\
    Nationality & Categorical & The nationality of student\\
    PlaceOfBirth & Categorical & The place of birth of student\\
    StageID & Categorical & Educational level (lower level, middle school, high school)\\
    GradeID & Categorical &  The grade of student\\
    SectionID & Categorical & The classroom (A, B, C)\\
    Topic & Categorical & Course topic (English, French, etc.)\\
    Semester & Categorical & School year semester (first, second)\\
    Relation & Categorical & Parent responsible for student (mom, father)\\
    Raisedhands & Numerical & How many times the student raises his/her hand\\
    VisitedResources & Numerical & How many times the student visits a course content\\
    AnnouncementsView & Numerical & How many times the student checks new announcements\\
    Discussion & Numerical & How many times the student participates in discussion\\
    ParentAnsweringSurvey & Numerical & Whether parent answered the surveys\\
    ParentschoolSatisfaction & Numerical & Whether the parents are satisfied\\
    StudentAbsenceDays & Numerical & The number of absence days\\
    Class (target variable) & Categorical & The grade’s level (low, middle, high)\\
    \bottomrule
    \end{tabular}
\end{table*}

\subsection{More Experimental Results}

In Table~\ref{tab:results_before}, we present the initial test accuracy and fairness measure before unlearning on OULAD, Student Performance, and xAPI-Edu-data datasets. As shown in the table, each dataset achieves strong test performance with different models under our training conditions. Following the unlearning process, our selective forgetting attacks have minimal impact on test performance. Recall that AEOD quantitatively measures the unfairness between two groups. The value domain of AEOD ranges from 0 to 1, with 0 standing for no discrimination and 1 indicating the maximum discrimination. As we can see, the AEOD values we consider are notably low for each dataset, suggesting relative fairness regarding the sensitive feature. However, our proposed selective forgetting attacks can effectively increase this fairness gap, thereby compromising fairness in the domain of educational data mining.

\begin{table*}[htbp]
\small
\centering
\caption{Evaluation of test accuracy and fairness measure before unlearning on OULAD, Student Performance, and xAPI-Edu-Data (sensitive feature: gender).}
\label{tab:results_before}
\begin{tabular}{cc|cc} 
\toprule
Dataset & Model & Test accuracy (\%) & AEOD\\ 
\midrule
\multirow{3}{*}{OULAD} & LR & $74.92 \pm 0.12$ & $0.1076 \pm 0.0011$\\
& MLP & $84.64 \pm 0.28$ & $0.0558 \pm 0.0040$\\
& MLP-2 & $84.74 \pm 0.30$ & $0.0453 \pm 0.0063$\\
\midrule
\multirow{3}{*}{Student Performance} & LR & $91.49 \pm 0.17$ & $0.0364 \pm 0.0101$\\
& MLP & $92.51 \pm 0.23$ & $0.0428 \pm 0.0019$\\
& MLP-2 & $91.03 \pm 0.44$ & $0.0347 \pm 0.0032$\\
\midrule
\multirow{3}{*}{xAPI-Edu-Data} & LR & $82.64 \pm 0.76$ & $0.0521 \pm 0.0061$\\
& MLP & $84.58 \pm 0.31$ & $0.0465 \pm 0.0041$\\
& MLP-2 & $82.29 \pm 0.13$ & $0.0407 \pm 0.0058$\\
\bottomrule
\end{tabular}
\end{table*}

\end{document}